\documentclass[conference,letterpaper,10pt]{IEEEtran}
\usepackage{fancyhdr,graphicx,times,amsmath,amssymb,cite,subfigure,color,url}
\usepackage[lined,ruled,commentsnumbered]{algorithm2e}
\allowdisplaybreaks

\fancypagestyle{plain}{
            \fancyhf{}         
            \fancyfoot[L]{}
            \fancyfoot[C]{}
            \fancyfoot[R]{}

}

{
            \fancyhf{}
            \fancyfoot[R]{}}

\begin{document}

\title{Offline EEG-Based Driver Drowsiness Estimation Using Enhanced Batch-Mode Active Learning (EBMAL) for Regression}

\author{\IEEEauthorblockN{Dongrui Wu\IEEEauthorrefmark{1}, Vernon J. Lawhern\IEEEauthorrefmark{2}\IEEEauthorrefmark{3}, Stephen Gordon\IEEEauthorrefmark{4}, Brent J. Lance\IEEEauthorrefmark{2}, Chin-Teng Lin\IEEEauthorrefmark{5}\IEEEauthorrefmark{6}}
\IEEEauthorblockA{\IEEEauthorrefmark{1}DataNova, NY USA}
\IEEEauthorblockA{\IEEEauthorrefmark{2}Human Research and Engineering Directorate, U.S. Army Research Laboratory, Aberdeen Proving Ground, MD USA}
\IEEEauthorblockA{\IEEEauthorrefmark{3}Department of Computer Science, University of Texas at San Antonio, San Antonio, TX USA}
\IEEEauthorblockA{\IEEEauthorrefmark{4}DCS Corp, Alexandria, VA USA}
\IEEEauthorblockA{\IEEEauthorrefmark{5}Brain Research Center, National Chiao-Tung University, Hsinchu, Taiwan}
\IEEEauthorblockA{\IEEEauthorrefmark{6}Faculty of Engineering and Information Technology, University of Technology, Sydney, Australia}
E-mail: drwu09@gmail.com, vernon.j.lawhern.civ@mail.mil, sgordon@dcscorp.com,\\ brent.j.lance.civ@mail.mil, ctlin@mail.nctu.edu.tw}
\IEEEoverridecommandlockouts
\IEEEpubid{\makebox[\columnwidth]{\hfill 978-1-5090-1897-0/16/\$31.00~\copyright2016 IEEE
                        } \hspace{\columnsep}\makebox[\columnwidth]{ }}
\maketitle

\begin{abstract}
There are many important regression problems in real-world brain-computer interface (BCI) applications, e.g., driver drowsiness estimation from EEG signals. This paper considers offline analysis: given a pool of unlabeled EEG epochs recorded during driving, how do we optimally select a small number of them to label so that an accurate regression model can be built from them to label the rest? Active learning is a promising solution to this problem, but interestingly, to our best knowledge, it has not been used for regression problems in BCI so far. This paper proposes a novel enhanced batch-mode active learning (EBMAL) approach for regression, which improves upon a baseline active learning algorithm by increasing the reliability, representativeness and diversity of the selected samples to achieve better regression performance. We validate its effectiveness using driver drowsiness estimation from EEG signals. However, EBMAL is a general approach that can also be applied to many other offline regression problems beyond BCI.
\end{abstract}


\section{Introduction}

EEG-based brain computer interfaces (BCIs) \cite{Muhl2014,Erp2012,Lance2012,Makeig2012,Wolpaw2012} have started to find real-world applications. However, usually a pilot session is required for each new subject to calibrate the BCI system, which negatively impacts its utility. It is very important to minimize this calibration effort, i.e., to achieve the best learning (classification, regression, or ranking) performance using as little subject-specific calibration data as possible.

There have been many approaches to reduce the BCI calibration effort. They can roughly be categorized into three groups: 1) methods to extract more robust and representative features, e.g., deep learning \cite{Hajinoroozi2015,Mao2014}, Riemannian geometry \cite{Barachant2012}, etc.; 2) methods to make use of axillary data from similar/relevant tasks, e.g., transfer learning/domain adaptation \cite{drwuaBCI2015,drwuACII2015,drwuSMC2015}, multi-task learning \cite{Alamgir2010}, etc.; and, 3) methods to optimize the calibration experiment design to generate or label more informative training data, e.g., active learning (AL) \cite{drwuRSVP2016,drwuSMC2015AL}, active class selection \cite{drwuACS2011}, etc.. It is interesting to note that these three groups are not mutually exclusive; in fact, methods in different groups can be combined for even better calibration performance. For example, active class selection and transfer learning were combined in \cite{drwuPLOS2013} to reduce the calibration effort in a virtual reality Stroop task, AL and transfer learning were combined in \cite{drwuSMC2014} for a visually evoked potential oddball task, and AL and domain adaptation were combined in \cite{drwuTNSRE2016} to reduce the calibration effort when switching between different EEG headsets.

This paper focuses on the third group, more specifically, AL to reduce offline BCI calibration effort, which considers the following problem: give a pool of unlabeled EEG epochs, how to optimally select a small number of them to label so that the learning (classification or regression) performance  can be maximized?

Considerable research has been done in this direction for classification problems in BCI \cite{drwuSMC2014,drwuTNSRE2016,drwuRSVP2016,drwuSMC2015AL,Chen2014,Moghadamfalahi2015,Zhao2011}, but to our best knowledge, AL has not been used for regression problems in BCI. In fact, compared with the extensive literature on AL for classification problems \cite{Settles2009}, AL for regression in general is significantly under-studied, not only for BCI. However, there are many interesting and challenging regression problems in BCI, e.g., driver drowsiness estimation from EEG signals \cite{drwuaBCI2015,Wei2015,Lin2009,Lin2006}. This is very important because according to the U.S. National Highway Traffic Safety Administration (NHTSA) \cite{NHTSA2011}, 2.5\% of fatal motor vehicle crashes between 2005 and 2009 (on average 886 annually in the U.S.) and 2.5\% of fatalities (on average 1,004 annually in the U.S.) involved drowsy driving. In our previous research we have focused on online driver drowsiness estimation from EEG signals \cite{drwuaBCI2015}. This paper considers offline analysis: given a pool of unlabeled EEG epochs recorded during driving, how do we optimally select a few to label so that an accurate regression model can be built from them to label the rest of the epoches?

This paper proposes a novel enhanced batch-mode active learning (EBMAL) approach for regression, which improves upon a baseline AL algorithm by increasing the reliability, representativeness and diversity of the selected samples to achieve better calibration performance. We use driver drowsiness estimation from EEG signals as an example to show that it significantly outperforms a baseline random sampling approach and two other AL approaches. However, our approach can also be applied to many other offline regression problems beyond BCI, e.g., estimating the continuous values of arousal, valence and dominance from speech signals \cite{drwuICME2010} in affective computing.

The remainder of this paper is organized as follows: Section~\ref{sect:EBMAL} introduces two baseline AL approaches and the proposed EBMAL approach to enhance them. Section~\ref{sect:experiment} describes the experiment setup and compares the performance of EBMAL with several other approaches. Section~\ref{sect:conclusions} draws conclusions.

\section{Enhanced Batch-Mode Active Learning (EBMAL)} \label{sect:EBMAL}

Our proposed EBMAL approach can be augmented to many existing AL algorithms to improve their performance. In this section we introduce two popular AL for regression approaches, point out their limitations, and show how they can be improved by EBMAL.

\subsection{AL for Regression by Query-by-Committee (QBC)} \label{sect:QBC}

Query-by-committee (QBC) is a very popular AL approach for both classification \cite{Abe1998,Seung1992,Freund1997,Settles2009} and regression \cite{Burbidge2007,Cohn1996,Krogh1995,RayChaudhuri1995,Settles2009,Demir2014} problems. Its basic idea is to build a committee of learners from existing labeled data (usually through bootstrapping), and then select the unlabeled samples on which the committee disagree the most to label.

More specifically, assume in a regression problem there are $N$ unlabeled samples $\{\mathbf{x}_n\}_{n=1}^N$, the committee consists of $P$ regression models, and the $p$th model's prediction for the $n$th unlabeled sample is $y_n^p$. Then, for each unlabeled sample, the QBC approach first computes the variance of the $P$ individual predictions, i.e. \cite{Burbidge2007},
\begin{align}
\sigma_n=\frac{1}{P}\sum_{p=1}^P\left(y_n^p-\bar{y}_n\right)^2, \quad n=1,...,N
\end{align}
where
\begin{align}
\bar{y}_n=\frac{1}{P}\sum_{p=1}^P y_n^p \label{eq:yn}
\end{align}
and then selects the top a few samples which have the maximal variance to label.

\subsection{AL for Regression by Expected Model Change Maximization (EMCM)} \label{sect:EMCM}

Expected model change maximization (EMCM) is also a very popular AL approach for classification \cite{Settles2009,Settles2008b,Settles2008,Cai2014}, ranking \cite{Donmez2008}, and regression \cite{Cai2013} problems. Cai et al. \cite{Cai2013} proposed an EMCM approach for both linear and nonlinear regression. In this subsection we introduce their linear approach, as only linear regression is considered in this paper.

Like in QBC, EMCM in \cite{Cai2013} also uses bootstrap to construct $P$ linear regression models. Assume the $p$th model's prediction for the $n$th unlabeled sample $\mathbf{x}_n$ is $y_n^p$. Then, for each unlabeled sample, it computes
\begin{align}
g(\mathbf{x}_n)=\frac{1}{P}\sum_{p=1}^P\left\| (y_n^p-\bar{y}_n)\mathbf{x}_n\right\|, \quad n=1,...,N
\end{align}
where $\bar{y}_n$ is again computed by (\ref{eq:yn}). EMCM finally selects the top a few samples which have the maximal $g(\mathbf{x}_n)$ to label.

\subsection{Limitations of the QBC and EMCM Approaches}

The above QBC and EMCM approaches, which will be called baseline AL approaches subsequently, have several limitations:
\begin{enumerate}
\item Usually the first batch of the samples for labeling are randomly selected, because the regression models cannot be constructed at the very beginning when no labeled data are available. However, there can still be better initialization approaches to select more reliable and representative seedling samples, without using any label information.
\item Sometimes the selected samples may be outliers, and hence labeling them not only waste the labeling effort, but may also deteriorates the regression performance. The baseline QBC and EMCM approaches do not have a mechanism to prevent outliers from being selected.
\item The baseline QBC and EMCM approaches consider each sample in the same batch independently, and no action is taken to reduce the redundancy among them, e.g., multiple selected samples in the same batch may be very close to each other, and hence using only one of them may be enough. The redundancy can be reduced by increasing the diversity of the samples in the same batch.
\end{enumerate}

\subsection{EBMAL} \label{sect:EBMAL}

In response to the above three limitations, we propose EBMAL in Algorithm~1, which employs the following three intuitive heuristics to improve the reliability, representativeness and diversity of samples selected by a baseline QBC or EMCM approach.

First, to select more reliable and representative seedling samples in the first batch, we perform $k$-means clustering on all unlabeled samples, where $k$ equals the batch size. We then compute the number of samples in each cluster to check if any cluster has a size no bigger than a certain empirical threshold, e.g., $\max(1,0.02N)$. If so, then the samples in that cluster are very likely to be outliers, and hence they are marked and restrained from being selected. We then perform $k$-means clustering again on the remaining unlabeled samples and repeat the check, until the number of samples in every cluster passes the size threshold. Then, for each cluster, we identify the sample that is closest to its centroid and select it for labeling. In this way we have selected $k$ samples in the initialization batch that are representative and diverse to label.

Second, to prevent potential outliers from being selected, we record all such samples from the initialization step and restrain them from consideration in all future iterations.

Third, in subsequent iterations after the initialization, instead of selecting directly the top $k$ unlabeled samples from a baseline AL approach, we now pre-select the top $2k$ samples using the baseline AL approach, and then perform $k$-means clustering on them, where $k$ again equals the batch size. This step partitions the $2k$ samples into $k$ groups according to their mutual distances. Then, for each cluster, we select the most informative sample (according to the baseline AL approach) for labelling. This ensures that the selected samples in the same batch are well-separated from each other, i.e., diversity is maintained.

\begin{algorithm}[h] 
\KwIn{$N$ unlabeled samples, $\{\mathbf{x}_n\}_{n=1}^N$\;
\hspace*{10mm} $k$, the batch size, which is also the number of clusters in $k$-means clustering\;
\hspace*{10mm} $M$, the number of batches\;
\hspace*{10mm} $\gamma$, determining the threshold for outlier identification}
\KwOut{The linear regression model $f(\mathbf{x})$.}
\For{$m=1,...,M$}{
\eIf{$m==1$}{
$S=\{\mathbf{x}_n\}_{n=1}^N$\;
$hasOutliers=True$\;
\While{$hasOutliers$}{
Perform $k$-means clustering on $S$ to obtain $k$ clusters, $C_i$, $i=1,...,k$\;
Set $p_i=|C_i|$\;
$hasOutliers=False$\;
\For{$i=1,...,k$}{
\If{$p_i\le \max(1,\gamma N)$}{
$S=S\setminus C_i$\;
$hasOutliers=True$\;}}}
\For{$i=1,...,k$}{
Select the sample closest to the centroid of $C_i$ to label\;}}
{Perform the baseline AL (e.g., QBC or EMCM) on $S$ and pre-select the top $2k$ most informative unlabeled samples\;
Perform $k$-means clustering on the $2k$ samples\;
\For{$i=1,...,k$}{
Select the most informative sample (according to the baseline AL) in Cluster $C_i$ to label\;}}}
Construct the linear regression model $f(\mathbf{x})$ from the $Mk$ labeled samples.
\caption{The EBMAL algorithm.} \label{alg:EAL}
\end{algorithm}

\section{Experiment and Results} \label{sect:experiment}

\subsection{Experiment Setup}

The experiment and data used in \cite{drwuaBCI2015} was again used in this study. We recruited 16 healthy subjects with normal/corrected-to-normal vision to participant in a sustained-attention driving experiment \cite{Chuang2012,Chuang2014}, consisting of a real vehicle mounted on a motion platform with 6 DOF immersed in a 360-degree virtual-reality (VR) scene. The Institutional Review Board of the Taipei Veterans General Hospital approved the experimental protocol, and each participant read and signed an informed consent form before the experiment began. Each experiment lasted for about 60-90 minutes and was conducted in the afternoon when the circadian rhythm of sleepiness reached its peak. To induce drowsiness during driving, the VR scenes simulated monotonous driving at a fixed speed (100 km/h) on a straight and empty highway. During the experiment, random lane-departure events were applied every 5-10 seconds, and participants were instructed to steer the vehicle to compensate for them immediately. The response time was recorded and later converted to a drowsiness index. Participants' scalp EEG signals were recorded using a 500Hz 32-channel Neuroscan system (30-channel EEGs plus 2-channel earlobes), and their cognitive states and driving performance were also monitored via a surveillance video camera and the vehicle trajectory throughout the experiment.

\subsection{Preprocessing and Feature Extraction}

The preprocessing and feature extraction methods were almost identical to those in our previous research \cite{drwuaBCI2015}, except that herein we used principal component features instead of the theta band power features for better regression performance.

The 16 subjects had different lengths of experiment, because the disturbances were presented randomly every 5-10 seconds. Data from one subject was not correctly recorded, so we used only 15 subjects. To ensure fair comparison, we used only the first 3,600 seconds data for each subject.

We defined a function \cite{Wei2015,drwuaBCI2015} to map the response time $\tau$ to a drowsiness index $y\in[0, 1]$:
\begin{align}
y=\max\left\{0,\,\frac{1-e^{-(\tau-\tau_0)}}{1+e^{-(\tau-\tau_0)}}\right\} \label{eq:y}
\end{align}
$\tau_0=1$ was used in this paper, as in \cite{drwuaBCI2015}. The drowsiness indices were then smoothed using a 90-second square moving-average window to reduce variations. This does not reduce the sensitivity of the drowsiness index because the cycle lengths of drowsiness fluctuations are longer than 4 minutes \cite{Makeig1993}.

We used EEGLAB \cite{Delorme2004} for EEG signal preprocessing. A 1-50 Hz band-pass filter was applied to remove high-frequency muscle artifacts, line-noise contamination and direct current drift. Next the EEG data were downsampled from 500 Hz to 250 Hz and re-referenced to averaged earlobes.

\begin{figure}[tbhp]\centering
\subfigure[]{\label{fig:drowy1}     \includegraphics[width=.46\linewidth,clip]{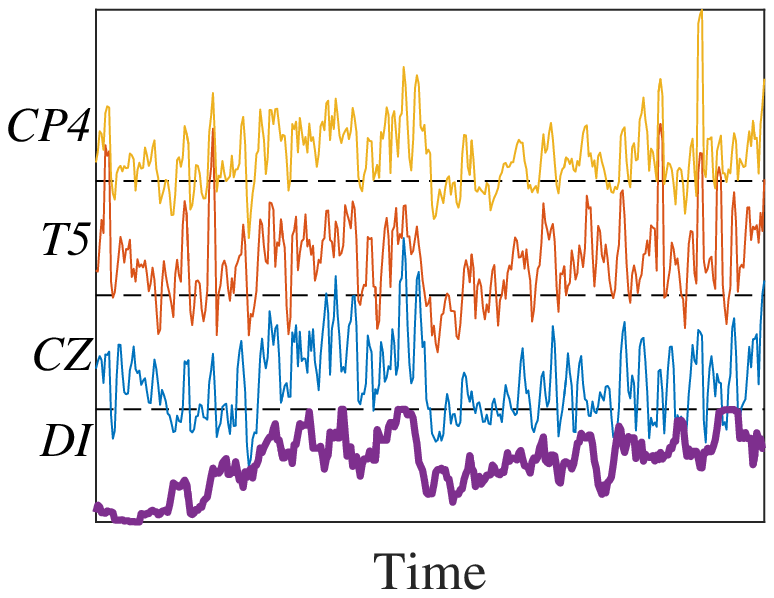}}
\subfigure[]{\label{fig:drowy2}     \includegraphics[width=.46\linewidth,clip]{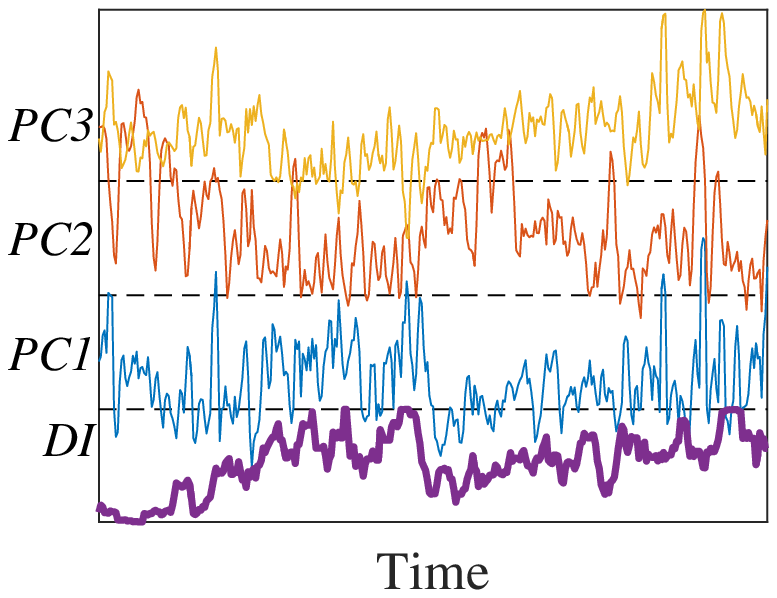}}
\caption{EEG features and the corresponding drowsiness index for Subject 1. (a) Theta band powers for three selected channels; (b) The top three principal component (PC) features.}
\end{figure}

We tried to predict the drowsiness index for each subject every 10 seconds, which is called a sample point in this paper. All 30 EEG channels were used in feature extraction. We epoched 30-second EEG signals right before each sample point, and computed the average power spectral density (PSD) in the theta band (4-7.5 Hz) for each channel using Welch's method \cite{Welch1967}, as research \cite{Makeig1996} has shown that theta band spectrum is a strong indicator of drowsiness. The theta band powers for three selected channels and the corresponding drowsiness index for a typical subject are shown in Fig.~\ref{fig:drowy1}. The correlation coefficients between the drowsiness index and CZ, T5 and CP5 theta band powers are 0.3005, 0.2706, and 0.3129, respectively, indicating considerable correlation.

Next, we converted the 30 theta band powers to dBs. To remove noises or bad channel readings, we removed channels whose maximum dBs were larger than 20. We then normalized the dBs of each remaining channel to mean zero and standard deviation one, and extracted a few (usually around 10) leading principal components, which accounted for 95\% of the variance. The projections of the dBs onto these principal components were then normalized to $[0, 1]$ and used as our features. Three such features for the same subject in Fig.~\ref{fig:drowy1} are shown in Fig.~\ref{fig:drowy2}. The correlation coefficients between the drowsiness index and the first three principal component scores are 0.2094, -0.6518 and 0.0169, respectively. Note that the maximum correlation is significantly improved by using the principal component features.

\subsection{Algorithms}

We compare the performances of five different sample selection strategies:
\begin{enumerate}
\item Baseline (BL), which randomly selects unlabeled samples for labelling.
\item QBC \cite{Burbidge2007}, which has been introduced in Section~\ref{sect:QBC}.
\item Enhanced QBC (EQBC), which is the QBC above enhanced by the EBMAL.
\item EMCM for linear regression \cite{Cai2013}, which has been introduced in Section~\ref{sect:EMCM}.
\item Enhanced EMCM (EEMCM), which is the EMCM above enhanced by the EBMAL.
\end{enumerate}
All five approaches build a linear ridge regression model from the labeled samples, as in \cite{drwuaBCI2015}. The ridge parameter $\sigma=0.01$ was used in all five algorithms.

\subsection{Evaluation Process and Performance Measures} \label{sect:process}

From the experiments we already knew the drowsiness indices for all $\sim$360 samples, obtained every 10 seconds from the first 3,600 seconds data. To evaluate the performances of different algorithms, for each subject, we first randomly selected 80\% of the $\sim$360 samples as our pool\footnote{For a fixed pool, EQBC and EEMCM give a deterministic selection sequence because there is no randomness involved. So, we need to vary the pool in order to study the statistical properties of EQBC and EEMCM. We did not use the traditional bootstrap approach, i.e., sampling with replacement to obtain the same number of samples as the original pool, because bootstrap introduces duplicate samples in the new pool, which does not happen in practice (a subject cannot have completely identical EEG responses at two different time instants), and also worsens the performances of QBC and EMCM (they may select multiple identical samples to label in the same batch).}, and then identified five samples to label in each batch by different algorithms, built a ridge regression model, and computed the root mean squared error (RMSE) and correlation coefficient (CC) as performance measures. The maximum number of samples to be labeled was fixed to be 60, corresponding to 12 batches.

We ran this evaluation process 30 times, each time with a randomly chosen 80\% population pool, to obtain statistically meaningful results.

\subsection{Experimental Results}

\begin{figure}[htpb]\centering
\subfigure[]{\label{fig:mRMSE}     \includegraphics[width=.46\linewidth,clip]{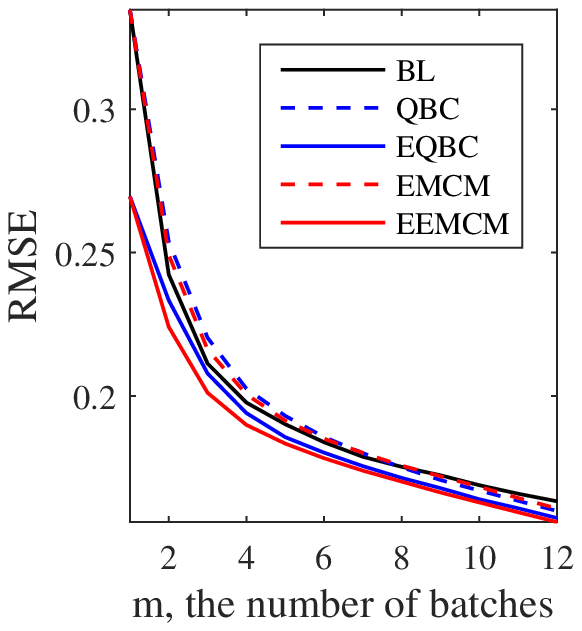}}
\subfigure[]{\label{fig:mCC}     \includegraphics[width=.46\linewidth,clip]{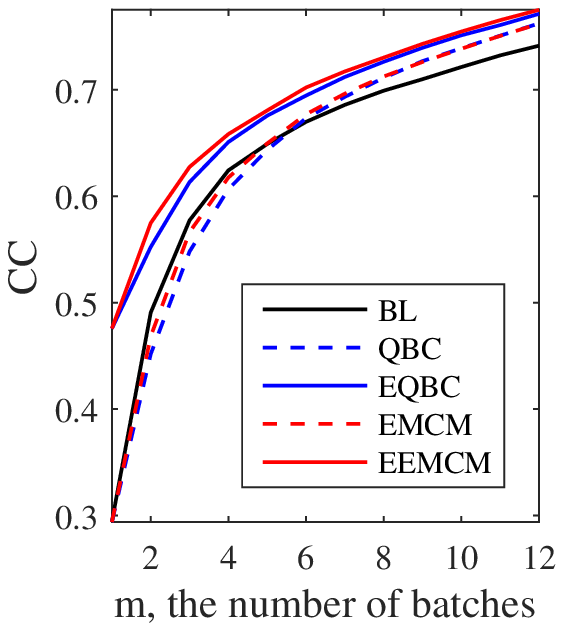}}
\caption{Average performances of the five algorithms across the 15 subjects. (a) RMSE; (b) CC.} \label{fig:mean}
\end{figure}

\begin{figure}[htpb]\centering
\subfigure[]{\label{fig:RMSEs}     \includegraphics[width=\linewidth,clip]{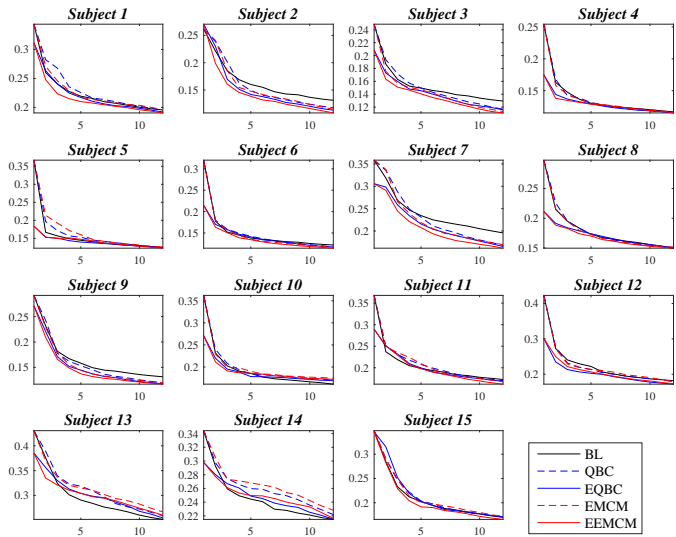}}
\subfigure[]{\label{fig:CCs}     \includegraphics[width=\linewidth,clip]{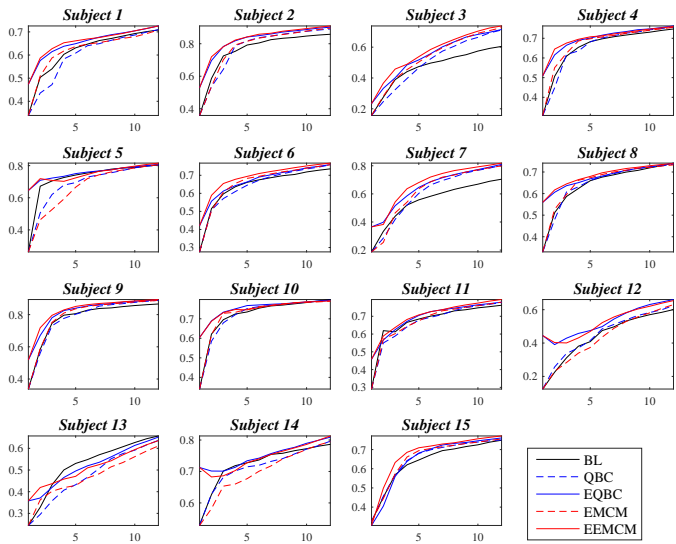}}
\caption{Performances of the five algorithms for each individual subject. Horizontal axis: $m$, the number of batches. (a) RMSE; (b) CC.} \label{fig:ind}
\end{figure}

The average RMSEs and CCs for the five algorithms across the 15 subjects are shown in Fig.~\ref{fig:mean}, and the RMSEs and CCs for the individual subjects are shown in Fig.~\ref{fig:ind}. Observe that all methods give better RMSEs and CCs as $m$ increases, which is intuitive. QBC and EMCM had very similar performance: both were comparable to or slightly worse than BL for small $m$, but as $m$ increased, they started to outperform BL. Remarkably, with the help of EBMAL, both EQBC and EEMCM outperformed the other three approaches for all $m$, although the performance improvement of EQBC and EEMCM over QBC and EMCM diminished as $m$ increased.

\begin{figure}[htpb]\centering
\subfigure[]{\label{fig:pmRMSE}     \includegraphics[width=.46\linewidth,clip]{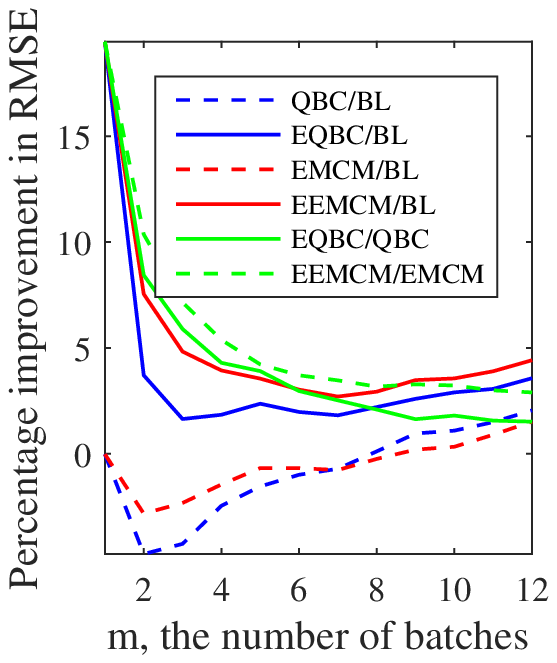}}
\subfigure[]{\label{fig:pmCC}     \includegraphics[width=.46\linewidth,clip]{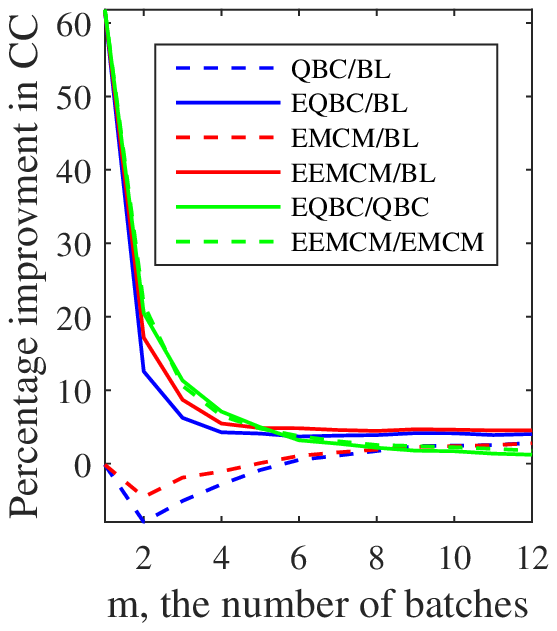}}
\caption{Percentage performance improvement between different pairs of algorithms. \emph{A/B} in the legend means the percentage performance improvement of Algorithm \emph{A} over Algorithm \emph{B}. (a) RMSE; (b) CC.} \label{fig:pmean}
\end{figure}

To better visualize the performance differences among different algorithms, in Fig.~\ref{fig:pmean} we plot the percentage performance improvement between different pairs of algorithms. Observe that although QBC and EMCM did not outperform BL for small $m$, the corresponding EQBC and EEMCM achieved the largest performance improvements over BL (and also QBC and EMCM) for small $m$, especially when $m=0$. This indicates that the new initialization strategy in EBMAL is indeed effective.

We also performed non-parametric multiple comparison tests using Dunn's procedure \cite{Dunn1961,Dunn1964} to determine if the differences between different pair of algorithms were statistically significant, with a $p$-value correction using the False Discovery Rate method \cite{Benjamini1995}. The $p$-values for RMSEs and CCs for different $m$ are shown in Tables~\ref{tab:Dunn} and \ref{tab:Dunn2}, respectively, where the statistically significant ones are marked in bold. Observe that QBC and EMCM had statistically significantly better RMSEs and CCs than BL for large $m$, but EQBC and EEMCM had statistically significantly better RMSEs and CCs for almost all $m$. The performance improvement of EQBC over QBC, and EEMCM over EMCM, was statistically significant for small $m$.

\begin{table}[htpb] \centering \setlength{\tabcolsep}{2mm}
\caption{$p$-values of non-parametric multiple comparisons on the RMSEs for different $m$.}   \label{tab:Dunn}
\begin{tabular}{c|cccccc}   \hline
   &           QBC &           EQBC  &            EMCM &          EEMCM & EQBC           &          EEMCM \\
$m$&    \emph{vs}  &      \emph{vs}  &       \emph{vs} &     \emph{vs}  & \emph{vs}      &      \emph{vs} \\
   &            BL &             BL  &              BL &             BL & QBC            &           EMCM \\ \hline
1  &         .5000 & \textbf{.0000}  &           .5000 & \textbf{.0000} & \textbf{.0000} & \textbf{.0000} \\
2  &         .1330 & \textbf{.0078}  &           .4294 & \textbf{.0032} & \textbf{.0002} & \textbf{.0043} \\
3  &         .1895 &          .0648  &           .4896 & \textbf{.0200} & \textbf{.0100} & \textbf{.0249} \\
4  &         .4157 &          .0763  &           .3026 & \textbf{.0094} & .0589          &          .0366 \\
5  &         .3761 &          .0652  &           .2203 & \textbf{.0067} & .1040          &          .0381 \\
6  &         .2022 &          .0296  &           .1483 & \textbf{.0013} & .1434          &          .0260  \\
7  &         .1464 & \textbf{.0236}  &           .1013 & \textbf{.0004} & .1474          &          .0267 \\
8  &         .1371 & \textbf{.0180}  &           .0668 & \textbf{.0003} & .1217          &          .0388 \\
9  &         .0565 & \textbf{.0128}  &           .0354 & \textbf{.0001} & .2063          &          .0410 \\
10 &\textbf{.0213} & \textbf{.0047}  &  \textbf{.0155} & \textbf{.0000} & .2828          &          .0490 \\
11 &\textbf{.0052} & \textbf{.0018}  &  \textbf{.0064} & \textbf{.0000} & .3371          &          .0533  \\
12 &\textbf{.0022} & \textbf{.0008}  &  \textbf{.0018} & \textbf{.0000} & .3970          &          .0852 \\
  \hline
\end{tabular}
\end{table}

\begin{table}[htpb] \centering \setlength{\tabcolsep}{2mm}
\caption{$p$-values of non-parametric multiple comparisons on the CCs for different $m$.}   \label{tab:Dunn2}
\begin{tabular}{c|cccccc}   \hline
   &            QBC &           EQBC &           EMCM &          EEMCM & EQBC          & EEMCM           \\
$m$&     \emph{vs}  &      \emph{vs} &      \emph{vs} &     \emph{vs}  & \emph{vs}     & \emph{vs}       \\
   &             BL &             BL &             BL &             BL & QBC           & EMCM            \\ \hline
1  &          .5000 & \textbf{.0000} &          .5000 & \textbf{.0000} & \textbf{.0000}& \textbf{.0000}  \\
2  &          .0814 & \textbf{.0003} &          .3963 & \textbf{.0032} & \textbf{.0000}& \textbf{.0000}  \\
3  &          .4090 & \textbf{.0041} &          .3842 & \textbf{.0059} & \textbf{.0038}& \textbf{.0170}  \\
4  &          .3035 & \textbf{.0020} &          .1289 & \textbf{.0004} & \textbf{.0079}& \textbf{.0096}  \\
5  &          .1515 & \textbf{.0022} &          .0436 & \textbf{.0002} & .0310         &          .0366  \\
6  &          .0661 & \textbf{.0003} & \textbf{.0175} & \textbf{.0000} & .0278         & \textbf{.0200}  \\
7  & \textbf{.0103} & \textbf{.0001} & \textbf{.0038} & \textbf{.0000} & .0762         & \textbf{.0114}  \\
8  & \textbf{.0061} & \textbf{.0000} & \textbf{.0018} & \textbf{.0000} & .0965         & \textbf{.0197}  \\
9  & \textbf{.0013} & \textbf{.0000} & \textbf{.0003} & \textbf{.0000} & .1687         &          .0251  \\
10 & \textbf{.0001} & \textbf{.0000} & \textbf{.0001} & \textbf{.0000} & .2663         &          .0290  \\
11 & \textbf{.0000} & \textbf{.0000} & \textbf{.0000} & \textbf{.0000} & .3231         &          .0255  \\
12 & \textbf{.0000} & \textbf{.0000} & \textbf{.0000} & \textbf{.0000} & .3511         &          .0407  \\
  \hline
\end{tabular}
\end{table}

\begin{figure}[htpb]\centering
\subfigure[]{\label{fig:mRMSEcomp}     \includegraphics[width=.46\linewidth,clip]{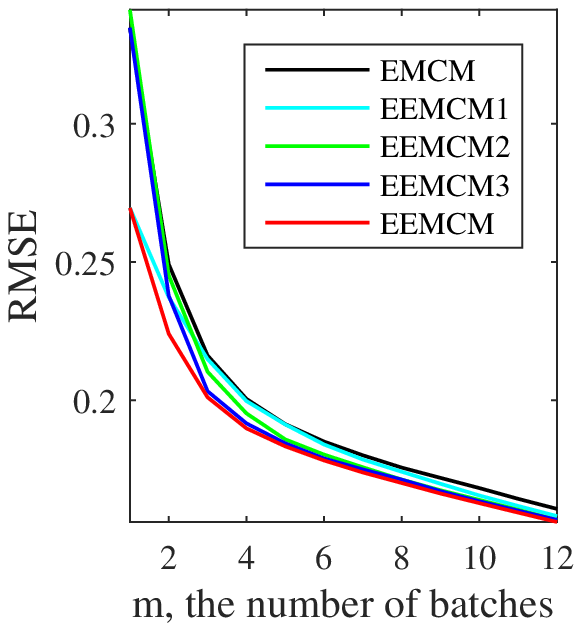}}
\subfigure[]{\label{fig:mCCcomp}     \includegraphics[width=.46\linewidth,clip]{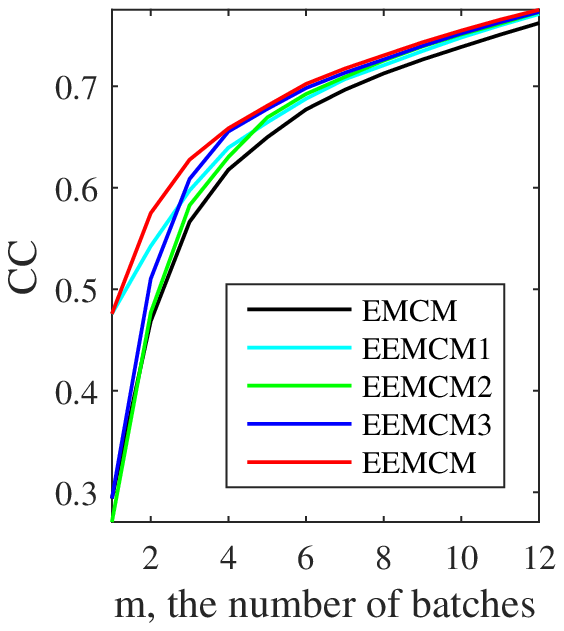}}
\caption{Effect of the individual enhancements in Section~\ref{sect:EBMAL}.} \label{fig:comp}
\end{figure}

It is also interesting to study if each of the three enhancements proposed in Section~\ref{sect:EBMAL} are necessary, and if so, what their individual effect is. For this purpose, we constructed three modified versions of the EBMAL algorithms: EBAML1, which employs only the first enhancement on more representative initialization; EBAML2, which employs only the second enhancement on better outlier handling; and, EBMAL3, which employs only the third enhancement on diversity. We then applied them to EMCM (the resulting algorithms are called EEMCM1, EEMCM2, and EEMCM3, respectively) and compared their performances with the baseline EMCM and the complement EEMCM. The results, averaged over 30 runs and 15 subjects, are shown in Fig.~\ref{fig:comp}. Observe that every enhancement outperformed the baseline EMCM. More specifically, the first enhancement on more representative initialization helped when $m$ was very small, especially at zero; the second and third enhancements on outlier handling and diversity helped when $m$ became larger. By combining the three enhancements, EEMCM achieved the best performance at both small and large $m$. This suggests that the three enhancements are complementary, and they are all essential to the improved performance of EBMAL.

In summary, we can conclude that our proposed EBMAL approach can significantly enhance a baseline AL for regression approach, especially when the number of labeled samples is very small (including zero). The three enhancements in EBMAL all contribute to its superior performance.

\section{Conclusions} \label{sect:conclusions}

Reducing the calibration data requirement in BCI systems is very important for their real-world applications. In our previous research we have extensively studied this in both online and offline BCI classification problems \cite{drwuSMC2014,drwuTNSRE2016,drwuRSVP2016,drwuSMC2015AL,drwuSMC2015,drwuACII2015}, and also online regression problems \cite{drwuaBCI2015}. This paper has proposed a novel EBMAL approach for offline BCI regression problems, and used EEG-based driver drowsiness estimation as an example to validate its performance. EBMAL solves the following problems: given a pool of unlabeled samples, how do we optimally select a small number of them to label so that an accurate regression model can be built from them to label the rest? Our proposed approach improves upon a baseline AL algorithm by increasing the reliability, representativeness and diversity of the selected samples to achieve better regression performance. To our best knowledge, this is the first time that active learning is used for regression problems in BCI. However, EBMAL is general and it can also be applied to many other offline regression problems beyond BCI, e.g., estimating the continuous values of arousal, valence and dominance from speech signals \cite{drwuICME2010} in affective computing.


\end{document}